\documentclass[11pt,a4paper]{article}
\usepackage[hyperref]{emnlp-ijcnlp-2019}
\usepackage{times}
\usepackage{latexsym}

\usepackage{url}

\aclfinalcopy 

\usepackage{makecell}
\makeatother

\title{Deep Bidirectional Transformers for Relation Extraction \\without Supervision}

\author{Yannis Papanikolaou, Ian Roberts, Andrea Pierleoni\\
    Healx, Cambridge, UK \\
  {\tt \{yannis.papanikolaou, ian.roberts, andrea.pierleoni\}@healx.io} \\}

\date{}

\begin{document}
\maketitle
\begin{abstract}
We present a novel framework to deal with relation extraction tasks in cases where there is complete lack of supervision, either in the form of gold annotations, or relations from a knowledge base. Our approach leverages syntactic parsing and pre-trained word embeddings to extract few but precise relations, which are then used to annotate a larger corpus, in a manner identical to distant supervision. The resulting data set is employed to fine tune a pre-trained BERT model in order to perform relation extraction. Empirical evaluation on four data sets from the biomedical domain shows that our method significantly outperforms two simple baselines for unsupervised relation extraction and, even if not using any supervision at all, achieves slightly worse results than the state-of-the-art in three out of four data sets. Importantly, we show that it is possible to successfully fine tune a large pre-trained language model with noisy data, as opposed to previous works that rely on gold data for fine tuning.

\end{abstract}

\section{Introduction}

The last years have seen a number of important advances in the field of Relation Extraction (RE), mainly based on deep learning models \citep{zeng2014relation, zeng2015distant, lin2016neural, zeng2016incorporating, wu2017adversarial, verga2018simultaneously}. These advances have led to significant improvements in benchmark tasks for RE. The above cases assume the existence of some form of supervision either manually annotated or distantly supervised data \citep{mintz2009distant}, where relations from a knowledge base are used in order to automatically annotate data, which then can be used as a noisy training set. For most real-world cases manually labeled data is either limited or completely missing, so typically one resorts to distant supervision to tackle a RE task.

There exist cases though, where even the distant supervision approach cannot be followed due to the lack of a knowledge base. This is often the case in domains like the Web or the biomedical literature, where entities of interest might be related with other entities and no available supervision signal exists.

In this work, we propose an approach to deal with such a scenario, from a purely unsupervised approach, that is without providing any manual annotation or any supervision whatsoever
. Our goal is to provide a framework that enables a pre-trained language model to be \textit{self-fine tuned}\footnote{We employ the term \textit{self-fine tuned} to denote that the model creates its own data set, without any supervision.} on a set of predefined relation types, in situations without any existing training data and without the possibility or budget for human supervision.

\begin{table}[!t]
    \centering
    \begin{tabular}{c|c|c}
    Verb & Relation & Similarity\\
    \hline
    \hline
    apply use &treat &0.40\\
investigate administer &treat &0.51\\
have manage &treat &0.60\\
evaluate improve &treat &0.41\\
be eradicate &treat &0.55\\
develop cause &cause &0.81\\
induce exacerbate &cause &0.58\\
know contribute &cause &0.41\\
result lead &cause &0.57\\
relate induce &cause &0.47\\
    \hline
    \hline
    \end{tabular}
    \caption{Examples of verb mappings for compound-disease relation. Each verb (can be n-gram as well) is mapped to its closest class (\textit{cause}, \textit{treat}) with pre-trained word embeddings.}
    \label{tbl:example_of_verb_mapping}
\end{table}

Our method proceeds as follows:
\begin{itemize}
    \item The data are first parsed syntactically, extracting relations of the form subject-verb-object. The resulting verbs are embedded in a vector space along with the relation types that we are interested to learn and each is mapped to their most similar relation type. Table \ref{tbl:example_of_verb_mapping} shows an example of this mapping process. This process is entirely automatic, we only provide the set of relation types that we are interested in and a threshold below which a verb is mapped to a \textit{Null} class.
    \item Subsequently, we use these extracted relations identically to a distant supervision signal to annotate automatically all co-occurrences of entities on a large corpus.
    \item The resulting data set is used to fine tune a Deep Bidirectional Transformer (BERT) model \citep{devlin2018bert}.
\end{itemize}

Importantly, the first step ensures that the resulting relations will have high precision (although at the expense of low recall), since they largely exclude the possibility of the two entities co-occurring randomly in the sentence, through the subject-verb-object association. In other words, we end up with a small, but high quality set of relations, which can then be used in a way identical to distant supervision.

The main contribution of this work is the introduction of a novel framework to deal with RE tasks without any supervision, either manually annotated data or known relations. Our approach is empirically evaluated on four data sets. A secondary implication of our work involves how we employ a pre-trained language model such as BERT: unlike previous approaches that employ a small gold data set, we show that it is possible to instead use a large noisy data set to successfully fine tune such a model.

The rest of the paper is organized as follows: we describe the related work in Section \ref{sec:related}, subsequently describing our method in Section \ref{sec:method} and presenting the empirical evaluation results in Section \ref{sec:experiments}. 

\section{Related work}
\label{sec:related}

Dealing with relation extraction in the absence of training data is not a novel task: for more than a decade, researchers have employed successfully techniques to tackle the lack of supervision, mainly by resorting to distant supervision \citep{mintz2009distant, riedel2010modeling}. This approach assumes the existence of a knowledge base, which contains already known relations between specific entities. These relations are then used to automatically annotate texts containing these entity pairs
. Although this approach leads to noisy labelling, it is cheap and has the ability to leverage a vast amount of training data. A great body of work has built upon this approach aiming to alleviate the noise in annotations, using formulations such as multi-label multi-instance learning \citep{surdeanu2012multi, zeng2015distant}, employing generative models to reduce wrong labelling \citep{takamatsu2012reducing}, developing different loss functions for relation extraction \citep{dos2015classifying, wang2016relation} or using side information to constraint predicted relations \citep{vashishth2018reside}.

More recently, a number of other interesting approaches have been presented aiming to deal with the lack of training data, with direct application to RE: data programming \citep{ratner2016data} provides a framework that allows domain experts to write labelling functions which are then denoised through a generative model. \citet{levy2017zero} have formulated the relation extraction task as a reading comprehension problem by associating one or more natural language questions with each relation. This approach enables generalization to unseen relations in a zero-shot setting.

Our work is different from the aforementioned approaches, in that it does not rely on the existence of any form of supervision. We build a model that is driven by the data, discovering a small set of precise relations, using them to annotate a larger corpus and being self-fine tuned to extract new relationships.

To train the RE classifier, we employ BERT, a recently proposed deep language model that achieved state-of-the-art results across a variety of tasks. BERT, similarly to the works of \citet{radford2018improving} and \citet{radford2018language}, builds upon the idea of pre-training a deep language model on massive amounts of data and then applies it (by fine tuning) to solve a diverse set of tasks. The building block of BERT is the Tranformer model \citep{vaswani2017attention}, a neural network cell that uses a multi-head, self-attention mechanism.

The first step of our approach is highly reminiscent of approaches from the open Information Extraction (openIE) literature \citep{banko2007open}. Indeed, similar to openIE approaches, we also use syntactic parsing to extract relations. Nevertheless, unlike openIE we are interested in a) specific types of entities which we assume that have been previously extracted with Named Entity Recognition (NER) and b) in specific, predefined types of relations between entities. We use syntactic parsing only as a means to extract a few precise relations and then follow an approach similar to distant supervision to train a neural relation extraction classifier. It should be noted though, that as a potential extension of this work we could employ more sophisticated techniques instead of syntactic parsing, similar to the latest openIE works \citep{yahya2014renoun}

\section{Method and Implementation Details}
\label{sec:method}
We present here the details of our method. First, we describe how we create our training set which results from a purely unsupervised procedure during which the only human intervention is to define the relation types of interest, e.g., 'treat' or 'associate'. Subsequently, we describe BERT, the model that we use in our approach.

\subsection{Training Set Creation}
\label{subsection:training_set_creation}
Our method assumes that the corpus is split in sentences\footnote{We can easily extend to cross-sentence relations, since the Transformer models which are the basis of BERT do not suffer from the problems encountered in LSTMs or CNNs for longer sequences, thanks to their self-attention mechanism.}, which are then passed through a NER model and a syntactic parser. We use the spaCy library\footnote{\url{https://spacy.io/}} for the above steps. 

Given a pair of two entities $A$ and $B$, we find their shortest dependency path and if one or more verbs $V$ are in that path we assume that $A$ is related to $B$ with $V$. The next step involves mapping the verbs to a set of predefined relation types, as shown in Table \ref{tbl:example_of_verb_mapping}. To do so, we embed both relation types and verbs to a continuous, lower-dimensional space with a pre-trained skip gram model \citep{mikolov2013distributed}, and map each verb to its closest relation type, if the cosine similarity of the two vectors is greater than a threshold (in initial small scale experiments using a validation set, we have found that a $threshold=0.4$ works well). Otherwise, the verb is not considered to represent a relation. In our experiments we used the pre-trained BioASQ word vectors\footnote{\url{ http://bioasq.lip6.fr/tools/BioASQword2vec/}}, since our relation extraction tasks come from the biomedical domain.

It is important to note that in the above procedure the only human involvement is defining the set of relation types that we are interested in. In that sense, this approach is neither domain or scale dependent: any set of relations can be used (coming from any domain) and likewise we can consider any number of relation types. 

The above procedure results in a small but relatively precise set of relations which can then be used in a way similar to distant supervision, to annotate all of our corpus. Nevertheless, there are a number of caveats to be taken into consideration:
\begin{itemize}
    \item As expected, there will be errors in the relations that come from the syntactic parsing and verbs mapping procedure.
    \item Our distant supervision-like approach comes also with inherent noise: we end up with a training set that has a lot of false negative and also a few false positive errors.
    \item The resulting training set will be largely imbalanced, since the way that we extract relations sacrifices recall for precision.
\end{itemize}

To deal with the above noise, we employ BERT as a relation extraction classifier. Furthermore, we use a balanced bagging approach to deal with class imbalance. Both approaches are described in detail in the following section.

\subsection{Deep Bidirectional Transformers}

BERT is a deep learning network that focus in learning general language representations which can then be used in downstream tasks. Much like the work of \citet{radford2018improving} and \citet{radford2018language}, the general idea is to leverage the expressive power of a deep Transformer architecture  that is pre-trained on a massive corpus on a language modelling task. Indeed, BERT comes in two flavors of 12 and 24 layers and 110M and 340M parameters respectively and is pre-trained on a concatenation of the English Wikipedia and the Book Corpus \citep{zhu2015aligning}. The resulting language model can then be fine tuned across a variety of different NLP tasks.

The main novelty of BERT is its ability to pre-train bidirectional representations by using a masked language model as a training objective. The idea behind the masked language model is to randomly mask some of the word tokens from the input, the objective being to predict what that word actually is, based on its context. The model is simultaneously trained on a second objective in order to model sentence relationships, that is, given two sentences $sent_a$ and $sent_b$ predict if $sent_b$ is the next sentence after $sent_a$. 

\begin{table*}[!t]
    \centering
    \begin{tabular}{c|ccccc}
    Data set & Relation (class) & \# Train (pos) & \# Dev (pos)& \# Test (pos)& \\
    \hline
    \hline
    Annotated&&&&\\
    \hline
    CDR     &Drug-Disease ($cause$)&3,596(1,453)&3,875(1,548)&3,805(1,482)\\
    GAD     &Disease-Gene ($cause$)&5,330(1,834)&-&-\\
    EUADR   &Disease-Gene ($cause$)&355(243)&-&-\\
    Healx CD &Drug-Disease ($treat$)&564(325)&-&-\\
    \hline
    Dist.Sup.&&&&\\
    \hline
    250k&Drug-Disease ($treat$)&250k(35k)&-&-\\
    full&Drug-Disease ($treat$)&8m(1.1m)&-&-\\
    \hline
    Our approach&&&&\\
    \hline
    250k&Drug-Disease ($treat$, $cause$)&250k(70k 10k)&-&-\\
    full&Drug-Disease ($treat$, $cause$)&8m(2.2m 325k)&-&-\\
    250k&Disease-Gene ($cause$)&250k(62k)&-&-\\
    full&Disease-Gene ($cause$)&9.1m(2.2m)&-&-\\
    \hline
    \hline
    \end{tabular}
    \caption{Data sets used in our experiments. 'Our approach' stands for the procedure described in  Section \ref{subsection:training_set_creation}. The Drug-Disease relation for our approach yields two positive classes, $treat$ and $cause$, therefore we report accordingly positives from each class in parentheses.}
    \label{tbl:dataset_stats}
\end{table*}

BERT has achieved state-of-the-art across eleven NLP tasks using the same pre-trained model and only with some simple fine tuning. This makes it particularly attractive for our use case, where we need a strong language model that will be able to learn from noisy patterns.

In order to further deal with the challenges mentioned in the previous section, in our experiments we fine tuned BERT for up to 5 epochs, since in early experiments we noticed that the model started overfitting to noise and validation loss started increasing after that point. 

\subsection{Balanced Bagging}
\label{subsection:bb}
In order to deal with class imbalance we employed balanced bagging \citep{tao2006asymmetric}, an ensembling technique where each component model is trained on a sub-sample of the data, such that the negative examples are roughly equal to the positive ones. To train each model of the ensemble, we sub-sample only the negative class so as to end up with a balanced set of positives and negatives. 

This sub-sampling of the negative class is important not only in order to alleviate the data imbalance, but also because the negative class will contain more noise than the positive by definition of our approach. In other words, since we consider as positives only a small set of relations coming from syntax parsing and verb mapping, it is more likely that a negative is in reality a positive sample rather than the opposite.

\section{Experiments}
\label{sec:experiments}
In this section we first describe the data sets used in experiments and the experimental setup and then present the results of our experiments.

\subsection{Data Sets and Setup}
We evaluate our method on four data sets coming from the biomedical domain, expressing disease-drug and disease-gene 
relations. Three of them are well known benchmark data sets for relation extraction: The Biocreative chemical-disease relations (CDR) data set \citep{li2016biocreative}, the Genetic Association Database (GAD) data set \citep{bravo2015extraction} and the EU-ADR data set \citep{van2012eu}. Additionally, we present a proprietary manually curated data set, Healx CD, expressing therapeutic drug-disease relations
. We consider only sentence-level relations, so we split CDR instances into sentences (the rest of the data sets are already at sentence-level). Statistics for the data sets are provided in Table \ref{tbl:dataset_stats}. We should note that for our approach we map each verb to the respective relation class that is depicted in Table \ref{tbl:dataset_stats} in parentheses.

As stated, we are mainly interested to understand how our proposed method performs under complete lack of training signal, so we compare it with two simple baselines for unsupervised relation extraction. The first, assumes that a sentence co-occurrence of two entities signals a positive relation, while the second is equivalent to the first two steps of our method, syntactic parsing followed by verb mapping to the relation types of interest. In other words, if two entities are connected in the shortest dependency path through a verb that is mapped to a class, they are considered to be related with that class.

Additionally, we would like to understand how our method performs against supervised methods, so for the first three data sets we compare it with a BERT model trained on the respective gold data, reporting also the current state-of-the-art, while for the Healx CD data set since there are no manual annotations, we compare our method against a distant supervision approach, retrieving ground truth relations from our internal knowledge base. 


Across all experiments and for all methods we use the same BERT model, BioBERT \citep{lee2019biobert}, which is a BERT model initialized with the model from \citet{devlin2018bert} and then pre-trained on PubMed, and thus more relevant to our tasks. That model is fine tuned on relation extraction classification using the code provided by the BioBERT authors, either on the gold or the distantly supervised or our approach's training set. We fine tune for up to 5 epochs with a learning rate of $0.00005$ and a batch size of $128$, keeping the model that achieves the best loss on the respective validation set. 

Finally, for the distant supervision as well as for our method, we use the previously mentioned balanced bagging approach, fine tuning an ensemble of ten models for each relation.

\subsection{Results}

\begin{table*}[!t]
    \centering
    \begin{tabular}{ccc|ccc}
    Data set&&Method & Precision & Recall & F1 \\
    \hline
    \hline
    CDR&&&&\\
    \hline
    &Unsupervised&Co-occurrences&30.9&100.0&47.2\\
    &&syntactic parsing+verb mapping&84.0&8.5&15.4\\
    &&Our method on BERT (250k)&49.4&76.3&60.4\\
    &&Our method on BERT (full)&50.1&81.3 &\textbf{62.2} \\
    \Xhline{0.00005\arrayrulewidth}
    &Supervised&SOTA \citep{verga2018simultaneously} & 64.2&68.5&66.3\\
    &&Gold Data on BERT &61.1&80.3&\textbf{70.4}\\
    \hline
    GAD&&&\\
    \hline
    &Unsupervised&Co-occurrences&34.4&100.0&51.2\\
    &&syntactic parsing+verb mapping&71.9&9.9&17.4\\
    &&Our method on BERT (250k)&53.1&82.8&64.6\\
    &&Our method on BERT (full)&56.9&90.1&\textbf{69.8}\\
    \Xhline{0.00005\arrayrulewidth}
    &Supervised&SOTA \citep{bhasuran2018automatic} &79.2&89.2&\textbf{83.9}\\
    &&Gold Data on BERT &76.4&87.7&81.7\\
    \hline
    EUADR&&&\\
    \hline
    &Unsupervised&Co-occurrences&68.5&100.0&\textbf{81.3}\\
    &&syntactic parsing+verb mapping&70.1&6.9&12.1\\
    &&Our method on BERT (250k)&71.7&79.4&75.5\\
    &&Our method on BERT (full)&75.5&87.9&81.2\\
    \Xhline{0.00005\arrayrulewidth}
    &Supervised&SOTA \citep{bhasuran2018automatic} &76.4&98.0&\textbf{85.3}\\
    &&Gold Data on BERT &78.0&93.9&85.2\\
    \hline
    Healx CD&& & \\
    \hline
    &Unsupervised&Co-occurrences&57.6&100.0&73.0\\
    &&syntactic parsing+verb mapping&91.0&17.9&29.9\\
    &&Our method on BERT (250k)&73.4&85.1&79.0\\
    &&Our method on BERT (full)&74.4&90.0&\textbf{81.4}\\
    \Xhline{0.00005\arrayrulewidth}
    &Supervised&Distant Supervision on BERT (250k) &83.3&83.1&83.4\\
    &&Distant Supervision on BERT (full)&87.1&83.2&\textbf{85.1}\\
    \hline
    \hline
    \end{tabular}
    \caption{Results on relation classification. State-of-the-art results were obtained from the corresponding papers. We averaged over five runs and report the evaluation metrics for a $0.5$ probability threshold.}
    \label{tab:sentence_classification}
\end{table*}

Table \ref{tab:sentence_classification} shows the results for the four data sets, reporting the average over five runs. For the GAD and EU-ADR data sets, we use the train and test splits provided by \citet{lee2019biobert}. Also, for CDR, since the state-of-the-art results \citep{verga2018simultaneously} are given at the abstract level, we re-run their proposed algorithm on our transformed sentence-level CDR data set, reporting results for a single model, without additional data (\citet{verga2018simultaneously} reports also results when adding weakly labeled data).

Let us first focus on the two unsupervised baselines. The first, dubbed 'co-occurrences', achieves a perfect recall since it considers all entity pairs co-occurrences as expressing a relation, but is clearly sub-optimal with regards to precision. The opposite behaviour is observed for the second baseline (syntactic parsing with verb mapping) since that one focuses in extracting high-precision relations, sacrificing recall: only entity pairs with a verb in between that is mapped to a relation are considered positives. Notably, this baseline achieves the highest precision in two out of four data sets, even compared to the supervised methods.

Our method proves significantly better compared to the other two unsupervised baselines, outperforming them by a large margin in all cases apart for EUADR. In that case our method is slightly worse than the co-occurrences baseline, since EUADR contains a big percentage of positives. Specifically, it is interesting to observe the improvement over the second baseline, which acts as a training signal for our method. Thanks to the predictive power and the robustness of BERT, our method manages to learn useful patterns from a noisy data set and actually improve substantially upon its training signal. 

An additional advantage of our method compared to the two other unsupervised baselines and similar approaches in general, is that it outputs a probability. Unlike the other methods, this probability allows us to tune our method for better precision or recall, depending on the application.

We then focus on comparing our proposed approach against the same BERT model fine tuned on supervised data, either manually annotated for the first three data sets, or distantly annotated for the fourth. For the first three data sets, we also report the current state-of-the-art results. Interestingly, even if our method is completely unsupervised, it is competitive with the state-of-the-art of fully supervised methods in three out of four cases, being inferior to them from 3.7 to 14.1 F1 points. On average, our method is worse by 7.5 F1 points against the best supervised model (either BERT or current state-of-the-art).

These results are particularly important, if we take into account that they come from a procedure that is fully unsupervised and which entails substantial noise from its sub-steps: the syntactic parsing may come with errors and mapping the verbs to relevant relation types is a process largely subject to the quality of the embeddings. Even worse, the relations obtained from the previous steps are used to automatically annotate all co-occurrences in a distant supervision-like fashion, which leads to even more noise. 

What we show empirically here is that despite all that noise coming from the above unsupervised procedure, we manage to successfully fine tune a deep learning model so as to achieve comparable performance to a fully supervised model. BERT is the main factor driving this robustness to noise and it can be mainly attributed to the fact that it consists of a very deep language model (112M parameters) and that it is pre-trained generatively on a massive corpus (3.3B words). The significance of these results is further amplified if we consider how scarce are labeled data for tasks such as relation extraction.

\subsection{Qualitative Analysis}

\begin{table*}[!t]
    \centering
    \begin{tabular}{ccccc}
    Sentence&class& BERT+gold&BERT+SP+VM&SP+VM\\
    \hline
    \hline
A patient with renal disease developed\\ coombs-positive DISEASE while\\receiving COMPOUND therapy.&cause&0.98&0.69&Null (developed)\\
    \hline
Five cases of DISEASE during treatment\\of loiasis with COMPOUND.&cause &  0.97&      0.95&       Null\\
    \hline
COMPOUND induced bradycardia in a\\patient with DISEASE.&Null&0.04&0.99&cause (induced)\\
    \hline
Neuroleptic drugs such as haloperidol,\\which block COMPOUND receptors,\\also cause DISEASE in rodents. & Null&0.92&0.99&treat (block)\\
    \hline
The results provide new insight\\into the potential role of ectopic\\hilar granule cells in the COMPOUND\\model of DISEASE.&cause&0.89&0.05&Null (provide)\\
    \hline
    \hline
    \end{tabular}
    \caption{Examples of predictions from the three methods on the $CDR$ data set. \textit{SP+VM} stands for the syntactic parsing+verb mapping baseline, while \textit{BERT+SP+VM} stands for our method. \textit{BERT+gold} is a BERT model trained on the gold $CDR$ training set. For \textit{SP+VM} we also provide the phrase verb in parentheses.}
    \label{tbl:examples_of_CDR_predictions}
\end{table*}

Although we showed empirically that our proposed approach is consistently capable to achieve results comparable to the SOTA, we would like to further focus on what are the weak points of the syntax parsing method and of our approach compared to a fully supervised approach.

To this end we inspected manually examples of predictions of the three aforementioned methods on the CDR data set, focusing on failures of our method and the syntactic parsing method which acts as training signal of our approach. Table \ref{tbl:examples_of_CDR_predictions} shows some characteristic cases:
\begin{itemize}
    \item In the first sentence, the syntactic parsing+verb mapping baseline (SP+VM) fails since the verb (\textit{developed}) is not associated with cause. Conversely our method, BERT with SP+VM manages to model correctly the sentence and extract the relation.
    \item SP+VM fails in the second example for the same reason, although the sentence is relatively simple.
    \item The third sentence represents also an interesting case, with SP+VM being "tricked" by the verb \textit{induced}. Our method also fails here, failing to attend correctly to the DISEASE masked entity.
    \item The fourth example represents a similar case, both BERT-based models are being tricked by the language. The SP+VM baseline is erroneously associating the verb \textit{block} to the relation \textit{treat} instead of cause.
    \item The fifth sentence resembles the first two: SP+VM fails to extract the relation for the same reason (verb in between). Our method fails too in that case, perhaps due to the relatively uncommon way that the causal relation is expressed (\textit{COMPOUND model of DISEASE}. 
\end{itemize}

While further inspecting the results, we also noticed a steady tendency of SP+VM to be able to capture relations in simpler (from a syntax perspective) and shorter sentences, while failing in the opposite case. 

Overall, we observe, as expected, that the SP+VM method is largely dependent on the simplicity of the expressed relation. Our method is clearly dependent on the quality of the syntax parsing, but manages up to a point to overcome low quality training data. To conclude, we can safely assume that our method would further benefit by replacing the SP+VM method with a more sophisticated unsupervised approach as the training signal, a future direction that we intend to take.


\section{Conclusions}
This work has introduced a novel framework to deal with relation extraction tasks in settings where there is complete lack of supervision. Our method employs syntactic parsing and word embeddings to extract a small set of precise relations which are then used to annotate a larger corpus, in the same way as distant supervision. With that data, we fine tune a pre-trained BERT model to perform relation extraction. 

We have empirically evaluated our method against two unsupervised baselines, a BERT model trained with gold or distantly supervised data and the current state-of-the-art. The results showed that our approach is significantly better than the unsupervised baselines, ranking slightly worse than the state-of-the-art in three out of four cases. 

Apart from presenting a novel perspective on how to train a relation extraction model in the absence of supervision, our work also shows empirically that it is possible to successfully fine tune a deep pre-trained language model with substantially noisy data.

We are interested in extending this paradigm to other areas of natural language processing tasks or adjusting our framework for more complex relation extraction tasks, as well as using more sophisticated unsupervised methods as training signal.

\subsection*{Acknowledgments}
We would like to thank Saatviga Sudhahar, for her insightful comments that greatly helped in improving this paper.

\bibliographystyle{acl_natbib}
\bibliography{main}

\begin{thebibliography}{29}
\expandafter\ifx\csname natexlab\endcsname\relax\def\natexlab#1{#1}\fi

\bibitem[{Banko et~al.(2007)Banko, Cafarella, Soderland, Broadhead, and
  Etzioni}]{banko2007open}
Michele Banko, Michael~J Cafarella, Stephen Soderland, Matthew Broadhead, and
  Oren Etzioni. 2007.
\newblock Open information extraction from the web.
\newblock In \emph{IJCAI}, pages 2670--2676.

\bibitem[{Bhasuran and Natarajan(2018)}]{bhasuran2018automatic}
Balu Bhasuran and Jeyakumar Natarajan. 2018.
\newblock Automatic extraction of gene-disease associations from literature
  using joint ensemble learning.
\newblock \emph{PloS one}, 13(7):e0200699.

\bibitem[{Bravo et~al.(2015)Bravo, Pi{\~n}ero, Queralt-Rosinach, Rautschka, and
  Furlong}]{bravo2015extraction}
{\`A}lex Bravo, Janet Pi{\~n}ero, N{\'u}ria Queralt-Rosinach, Michael
  Rautschka, and Laura~I Furlong. 2015.
\newblock Extraction of relations between genes and diseases from text and
  large-scale data analysis: implications for translational research.
\newblock \emph{BMC bioinformatics}, 16(1):55.

\bibitem[{Devlin et~al.(2018)Devlin, Chang, Lee, and
  Toutanova}]{devlin2018bert}
Jacob Devlin, Ming-Wei Chang, Kenton Lee, and Kristina Toutanova. 2018.
\newblock Bert: Pre-training of deep bidirectional transformers for language
  understanding.
\newblock \emph{arXiv preprint arXiv:1810.04805}.

\bibitem[{Lee et~al.(2019)Lee, Yoon, Kim, Kim, Kim, So, and
  Kang}]{lee2019biobert}
Jinhyuk Lee, Wonjin Yoon, Sungdong Kim, Donghyeon Kim, Sunkyu Kim, Chan~Ho So,
  and Jaewoo Kang. 2019.
\newblock Biobert: pre-trained biomedical language representation model for
  biomedical text mining.
\newblock \emph{arXiv preprint arXiv:1901.08746}.

\bibitem[{Levy et~al.(2017)Levy, Seo, Choi, and Zettlemoyer}]{levy2017zero}
Omer Levy, Minjoon Seo, Eunsol Choi, and Luke Zettlemoyer. 2017.
\newblock Zero-shot relation extraction via reading comprehension.
\newblock \emph{arXiv preprint arXiv:1706.04115}.

\bibitem[{Li et~al.(2016)Li, Sun, Johnson, Sciaky, Wei, Leaman, Davis,
  Mattingly, Wiegers, and Lu}]{li2016biocreative}
Jiao Li, Yueping Sun, Robin~J Johnson, Daniela Sciaky, Chih-Hsuan Wei, Robert
  Leaman, Allan~Peter Davis, Carolyn~J Mattingly, Thomas~C Wiegers, and Zhiyong
  Lu. 2016.
\newblock Biocreative v cdr task corpus: a resource for chemical disease
  relation extraction.
\newblock \emph{Database}, 2016.

\bibitem[{Lin et~al.(2016)Lin, Shen, Liu, Luan, and Sun}]{lin2016neural}
Yankai Lin, Shiqi Shen, Zhiyuan Liu, Huanbo Luan, and Maosong Sun. 2016.
\newblock Neural relation extraction with selective attention over instances.
\newblock In \emph{Proceedings of the 54th Annual Meeting of the Association
  for Computational Linguistics (Volume 1: Long Papers)}, volume~1, pages
  2124--2133.

\bibitem[{Mikolov et~al.(2013)Mikolov, Sutskever, Chen, Corrado, and
  Dean}]{mikolov2013distributed}
Tomas Mikolov, Ilya Sutskever, Kai Chen, Greg~S Corrado, and Jeff Dean. 2013.
\newblock Distributed representations of words and phrases and their
  compositionality.
\newblock In \emph{Advances in neural information processing systems}, pages
  3111--3119.

\bibitem[{Mintz et~al.(2009)Mintz, Bills, Snow, and
  Jurafsky}]{mintz2009distant}
Mike Mintz, Steven Bills, Rion Snow, and Dan Jurafsky. 2009.
\newblock Distant supervision for relation extraction without labeled data.
\newblock In \emph{Proceedings of the Joint Conference of the 47th Annual
  Meeting of the ACL and the 4th International Joint Conference on Natural
  Language Processing of the AFNLP: Volume 2-Volume 2}, pages 1003--1011.
  Association for Computational Linguistics.

\bibitem[{Radford et~al.(2018{\natexlab{a}})Radford, Narasimhan, Salimans, and
  Sutskever}]{radford2018improving}
Alec Radford, Karthik Narasimhan, Tim Salimans, and Ilya Sutskever.
  2018{\natexlab{a}}.
\newblock Improving language understanding by generative pre-training.
\newblock Technical report, Technical report, OpenAi.

\bibitem[{Radford et~al.(2018{\natexlab{b}})Radford, Wu, Child, Luan, Amodei,
  and Sutskever}]{radford2018language}
Alec Radford, Jeffrey Wu, Rewon Child, David Luan, Dario Amodei, and Ilya
  Sutskever. 2018{\natexlab{b}}.
\newblock Language models are unsupervised multitask learners.
\newblock Technical report, Technical report, OpenAi.

\bibitem[{Ratner et~al.(2016)Ratner, De~Sa, Wu, Selsam, and
  R{\'e}}]{ratner2016data}
Alexander~J Ratner, Christopher~M De~Sa, Sen Wu, Daniel Selsam, and Christopher
  R{\'e}. 2016.
\newblock Data programming: Creating large training sets, quickly.
\newblock In \emph{Advances in neural information processing systems}, pages
  3567--3575.

\bibitem[{Riedel et~al.(2010)Riedel, Yao, and McCallum}]{riedel2010modeling}
Sebastian Riedel, Limin Yao, and Andrew McCallum. 2010.
\newblock Modeling relations and their mentions without labeled text.
\newblock In \emph{Joint European Conference on Machine Learning and Knowledge
  Discovery in Databases}, pages 148--163. Springer.

\bibitem[{dos Santos et~al.(2015)dos Santos, Xiang, and
  Zhou}]{dos2015classifying}
Cicero dos Santos, Bing Xiang, and Bowen Zhou. 2015.
\newblock Classifying relations by ranking with convolutional neural networks.
\newblock In \emph{Proceedings of the 53rd Annual Meeting of the Association
  for Computational Linguistics and the 7th International Joint Conference on
  Natural Language Processing (Volume 1: Long Papers)}, volume~1, pages
  626--634.

\bibitem[{Surdeanu et~al.(2012)Surdeanu, Tibshirani, Nallapati, and
  Manning}]{surdeanu2012multi}
Mihai Surdeanu, Julie Tibshirani, Ramesh Nallapati, and Christopher~D Manning.
  2012.
\newblock Multi-instance multi-label learning for relation extraction.
\newblock In \emph{Proceedings of the 2012 joint conference on empirical
  methods in natural language processing and computational natural language
  learning}, pages 455--465. Association for Computational Linguistics.

\bibitem[{Takamatsu et~al.(2012)Takamatsu, Sato, and
  Nakagawa}]{takamatsu2012reducing}
Shingo Takamatsu, Issei Sato, and Hiroshi Nakagawa. 2012.
\newblock Reducing wrong labels in distant supervision for relation extraction.
\newblock In \emph{Proceedings of the 50th Annual Meeting of the Association
  for Computational Linguistics: Long Papers-Volume 1}, pages 721--729.
  Association for Computational Linguistics.

\bibitem[{Tao et~al.(2006)Tao, Tang, Li, and Wu}]{tao2006asymmetric}
Dacheng Tao, Xiaoou Tang, Xuelong Li, and Xindong Wu. 2006.
\newblock Asymmetric bagging and random subspace for support vector
  machines-based relevance feedback in image retrieval.
\newblock \emph{IEEE Transactions on Pattern Analysis \& Machine Intelligence},
  (7):1088--1099.

\bibitem[{Van~Mulligen et~al.(2012)Van~Mulligen, Fourrier-Reglat, Gurwitz,
  Molokhia, Nieto, Trifiro, Kors, and Furlong}]{van2012eu}
Erik~M Van~Mulligen, Annie Fourrier-Reglat, David Gurwitz, Mariam Molokhia,
  Ainhoa Nieto, Gianluca Trifiro, Jan~A Kors, and Laura~I Furlong. 2012.
\newblock The eu-adr corpus: annotated drugs, diseases, targets, and their
  relationships.
\newblock \emph{Journal of biomedical informatics}, 45(5):879--884.

\bibitem[{Vashishth et~al.(2018)Vashishth, Joshi, Prayaga, Bhattacharyya, and
  Talukdar}]{vashishth2018reside}
Shikhar Vashishth, Rishabh Joshi, Sai~Suman Prayaga, Chiranjib Bhattacharyya,
  and Partha Talukdar. 2018.
\newblock Reside: Improving distantly-supervised neural relation extraction
  using side information.
\newblock In \emph{Proceedings of the 2018 Conference on Empirical Methods in
  Natural Language Processing}, pages 1257--1266.

\bibitem[{Vaswani et~al.(2017)Vaswani, Shazeer, Parmar, Uszkoreit, Jones,
  Gomez, Kaiser, and Polosukhin}]{vaswani2017attention}
Ashish Vaswani, Noam Shazeer, Niki Parmar, Jakob Uszkoreit, Llion Jones,
  Aidan~N Gomez, {\L}ukasz Kaiser, and Illia Polosukhin. 2017.
\newblock Attention is all you need.
\newblock In \emph{Advances in neural information processing systems}, pages
  5998--6008.

\bibitem[{Verga et~al.(2018)Verga, Strubell, and
  McCallum}]{verga2018simultaneously}
Patrick Verga, Emma Strubell, and Andrew McCallum. 2018.
\newblock Simultaneously self-attending to all mentions for full-abstract
  biological relation extraction.
\newblock In \emph{Proceedings of the 2018 Conference of the North American
  Chapter of the Association for Computational Linguistics: Human Language
  Technologies, Volume 1 (Long Papers)}, volume~1, pages 872--884.

\bibitem[{Wang et~al.(2016)Wang, Cao, de~Melo, and Liu}]{wang2016relation}
Linlin Wang, Zhu Cao, Gerard de~Melo, and Zhiyuan Liu. 2016.
\newblock Relation classification via multi-level attention cnns.
\newblock In \emph{Proceedings of the 54th Annual Meeting of the Association
  for Computational Linguistics (Volume 1: Long Papers)}, volume~1, pages
  1298--1307.

\bibitem[{Wu et~al.(2017)Wu, Bamman, and Russell}]{wu2017adversarial}
Yi~Wu, David Bamman, and Stuart Russell. 2017.
\newblock Adversarial training for relation extraction.
\newblock In \emph{Proceedings of the 2017 Conference on Empirical Methods in
  Natural Language Processing}, pages 1778--1783.

\bibitem[{Yahya et~al.(2014)Yahya, Whang, Gupta, and Halevy}]{yahya2014renoun}
Mohamed Yahya, Steven Whang, Rahul Gupta, and Alon Halevy. 2014.
\newblock Renoun: Fact extraction for nominal attributes.
\newblock In \emph{Proceedings of the 2014 Conference on Empirical Methods in
  Natural Language Processing (EMNLP)}, pages 325--335.

\bibitem[{Zeng et~al.(2015)Zeng, Liu, Chen, and Zhao}]{zeng2015distant}
Daojian Zeng, Kang Liu, Yubo Chen, and Jun Zhao. 2015.
\newblock Distant supervision for relation extraction via piecewise
  convolutional neural networks.
\newblock In \emph{Proceedings of the 2015 Conference on Empirical Methods in
  Natural Language Processing}, pages 1753--1762.

\bibitem[{Zeng et~al.(2014)Zeng, Liu, Lai, Zhou, Zhao
  et~al.}]{zeng2014relation}
Daojian Zeng, Kang Liu, Siwei Lai, Guangyou Zhou, Jun Zhao, et~al. 2014.
\newblock Relation classification via convolutional deep neural network.

\bibitem[{Zeng et~al.(2016)Zeng, Lin, Liu, and Sun}]{zeng2016incorporating}
Wenyuan Zeng, Yankai Lin, Zhiyuan Liu, and Maosong Sun. 2016.
\newblock Incorporating relation paths in neural relation extraction.
\newblock \emph{arXiv preprint arXiv:1609.07479}.

\bibitem[{Zhu et~al.(2015)Zhu, Kiros, Zemel, Salakhutdinov, Urtasun, Torralba,
  and Fidler}]{zhu2015aligning}
Yukun Zhu, Ryan Kiros, Rich Zemel, Ruslan Salakhutdinov, Raquel Urtasun,
  Antonio Torralba, and Sanja Fidler. 2015.
\newblock Aligning books and movies: Towards story-like visual explanations by
  watching movies and reading books.
\newblock In \emph{Proceedings of the IEEE international conference on computer
  vision}, pages 19--27.

\end{thebibliography}
\end{document}